\documentclass[letterpaper, journal]{IEEEtran}

\usepackage{mathtools}    
\usepackage{amssymb}
\usepackage{amsfonts}

\usepackage{graphicx}
\usepackage[dvipsnames,table]{xcolor}

\usepackage{cite}
\usepackage{url}
\usepackage{colortbl}
\usepackage{threeparttable}
\usepackage{booktabs}
\usepackage{multirow}
\usepackage{array}
\usepackage{tabularx}
\usepackage{placeins}
\usepackage{makecell}


\usepackage{siunitx}
\newcolumntype{Y}{>{\centering\arraybackslash}X}

\usepackage[caption=false,font=footnotesize]{subfig}

\usepackage{algorithm}
\usepackage{algorithmic}

\usepackage{textcomp}
\usepackage{verbatim}
\usepackage[utf8]{inputenc}
\usepackage{balance}
\usepackage{float}
\usepackage{listings}
\usepackage[normalem]{ulem}
\usepackage{pifont}
\usepackage{tcolorbox}
\tcbuselibrary{listings,breakable,skins}
\usepackage{lipsum}
\usepackage{longtable}
\usepackage{seqsplit}
\newif\ifshowcomment
\newif\ifnumberrevision
\newif\ifcolorrevision
\newif\ifstrikeremovel

\showcommenttrue
\numberrevisiontrue
\colorrevisiontrue
\strikeremoveltrue

\definecolor{fig07cOrange}{HTML}{DC843F}
\definecolor{fig07cGreen}{HTML}{3D9970}
\definecolor{fig07cPink}{HTML}{E377A0}
\definecolor{fig07cBlue}{HTML}{074B90}
\definecolor{fig07cGold}{HTML}{C9A020}
\definecolor{fig07cPurple}{HTML}{7B52AB}

\title{Online Multimodal Workload Assessment in Contact-Rich Physical Human-Robot Interaction}

\author{
Yanyi Chen\textsuperscript{1},
Fan Yang\textsuperscript{2},
and Min Deng\textsuperscript{1,*}
\thanks{\textsuperscript{*}Corresponding author: Min Deng.}
\thanks{\textsuperscript{1}Yanyi Chen is with the Department of Civil and Environmental Engineering, University of Tennessee, Knoxville, TN 37996, USA (e-mail: yanychen@utk.edu).}
\thanks{\textsuperscript{2}Fan Yang is with the College of Information and Communications, University of South Carolina, SC 29208, USA (e-mail: YANG259@mailbox.sc.edu).}
\thanks{\textsuperscript{1}Min Deng is with the Department of Civil and Environmental Engineering, University of Tennessee, Knoxville, TN 37996, USA (e-mail: mindeng@utk.edu).}
}

\begin{document}
\maketitle

\begin{abstract}
Contact-rich physical human--robot interaction (pHRI) imposes time-varying demands associated with physical interaction, motor regulation, and physiological response, motivating continuous assessment of interaction workload. This paper presents an online multimodal assessment framework that integrates interaction wrench, planar tool-center-point (TCP) kinematics, and skin conductance level (SCL) into four interpretable workload-related factors. Their relative contributions are adjusted using path curvature to reflect changes in motion demand and task progression to account for gradual physiological variation over time. The framework was evaluated with 24 participants across 18 controlled combinations of temperature, acoustic noise, and illuminance under two admittance-control modes. Strict leave-one-subject-out (LOSO) evaluation used standardized pupil diameter ($\mathrm{PD}_z$) as an independent physiological reference and included comparisons with static variants and representative state-of-the-art learning-based baselines. The proposed framework achieves a cohort-mean $30\,\mathrm{s}$ block-wise Spearman correlation of $\rho_{30}=0.308$ with the physiological reference, with positive subject-level correspondence in 23 of 24 participants. Its overall performance is comparable to the state-of-the-art learning-based baseline.  At the same time, our framework keeps the assessment process transparent through explicit workload-related factors and defined weighting rules, while outperforming the corresponding fixed-weight formulation. The framework also maintains consistent performance across the two tested admittance-control modes. These results support a transparent and interpretable approach to continuous interaction workload assessment in contact-rich pHRI.

\end{abstract}

\section{Introduction}
\label{sec:introduction}

\IEEEPARstart{R}{obotic} systems are increasingly extending beyond isolated automation~\cite{he2026constraintgroundedreinforcementlearningvariable} toward close interaction with human operators in domains such as advanced manufacturing~\cite{polish2025}, intelligent construction~\cite{deng2026integrating, chen2026perception}, and robot-assisted surgery~\cite{fu2025}. In contact-rich physical human--robot interaction (pHRI), humans and robots jointly manipulate tools or objects while continuously regulating motion and contact forces during tasks such as cooperative payload transportation~\cite{co-carrying2024framework}, robotic surface polishing~\cite{polish2025}, and high-frequency ultrasound scanning~\cite{see2025ultrasound, force2024ultrasound}. Some existing pHRI research focuses on operator workload monitoring to emphasize safe and compliant physical interaction. Operator workload in pHRI has been studied from both physical and cognitive perspectives. Physical workload and ergonomic demand are assessed using interaction forces, kinematics, electromyography (EMG), and posture~\cite{kiki2025estimating, peternel2018robot}. Cognitive workload and related physiological responses are assessed using electroencephalography (EEG), functional near-infrared spectroscopy (fNIRS), heart-rate variability (HRV), Electrodermal activity (EDA), and pupillometry~\cite{progress2023review, belkaidMutualGazeRobot2021, lorenzini2022ergonomic, storm2022, upasani2023eye, aygunetal22sensors}. 

From raw data to state estimation, a range of state-of-the-art learning methods has been developed for both single-modal and multimodal measurements. Support vector machines (SVMs) and random forests (RFs)~\cite{cakit2025}, multilayer perceptrons (MLPs)~\cite{aygunetal22sensors}, temporal convolutional networks (TCNs)~\cite{xia2025large}, and long short-term memory (LSTM) networks~\cite{asgher2020enhanced} can model nonlinear relationships between measured signals and workload-related references. However, multimodal fusion often combines signals from similar information sources. Physical interaction behavior and physiological response are still modeled separately rather than integrated into a unified workload estimate. Learning-based models also provide limited insight into how individual information sources contribute to the estimate.

Motivated by these gaps, we propose an online, deployable, and interpretable framework for assessing the overall workload experienced during contact-rich pHRI using multimodel data. We refer to this overall workload as \textit{interaction workload} to represent the time-varying workload experienced by the operator, which is jointly reflected in interaction behavior and physiological response. The framework derives four interpretable workload-related factors from interaction wrenches, planar tool-center-point (TCP) motion, and Galvanic Skin Response (GSR). And they are adjusted their relative contributions according to path curvature and task progression. Furthermore, it is evaluated in a contact-rich path-tracing task across 18 controlled combinations of temperature, acoustic noise, and illuminance designed to reflect environmental variability in industrial workplaces, under two admittance-control modes. A strict leave-one-subject-out (LOSO) protocol is used to evaluate the proposed framework against its static and contextual variants as well as MLP, TCN, and LSTM learning-based baselines. Standardized pupil diameter, $\mathrm{PD}_z$, serves as an independent physiological reference for offline validation. The proposed framework achieves a cohort-mean $30\,\mathrm{s}$ block-wise Spearman correlation of $\rho_{30}=0.308$ with the physiological reference, with positive correspondence in 23 of 24 participants. Its performance is comparable to the strongest learning-based baseline, MLP ($\rho_{30}=0.313$), and remains consistent across the two control modes.

The main contributions of this work are as follows:
\begin{itemize}

\item We propose an online multimodal framework for assessing interaction workload in contact-rich pHRI by jointly using interaction wrench, TCP kinematics, and skin conductance level (SCL) to capture complementary physical-interaction and physiological responses.

\item We formulate four interpretable workload-related factors for corrective interaction, contact regulation, kinematic irregularity, and autonomic arousal, and introduce context-adaptive weighting based on path curvature and task progression.

\item We validate the framework using strict LOSO evaluation across 24 participants, 18 environmental conditions, and two admittance-control modes, with comparisons against framework variants and representative state-of-the-art learning-based baselines using an independent physiological reference.

\end{itemize}
\section{Interaction Workload Assessment Framework}
\label{sec:online_iw}

\begin{figure*}[!t]
\centering
\includegraphics[width=\textwidth]{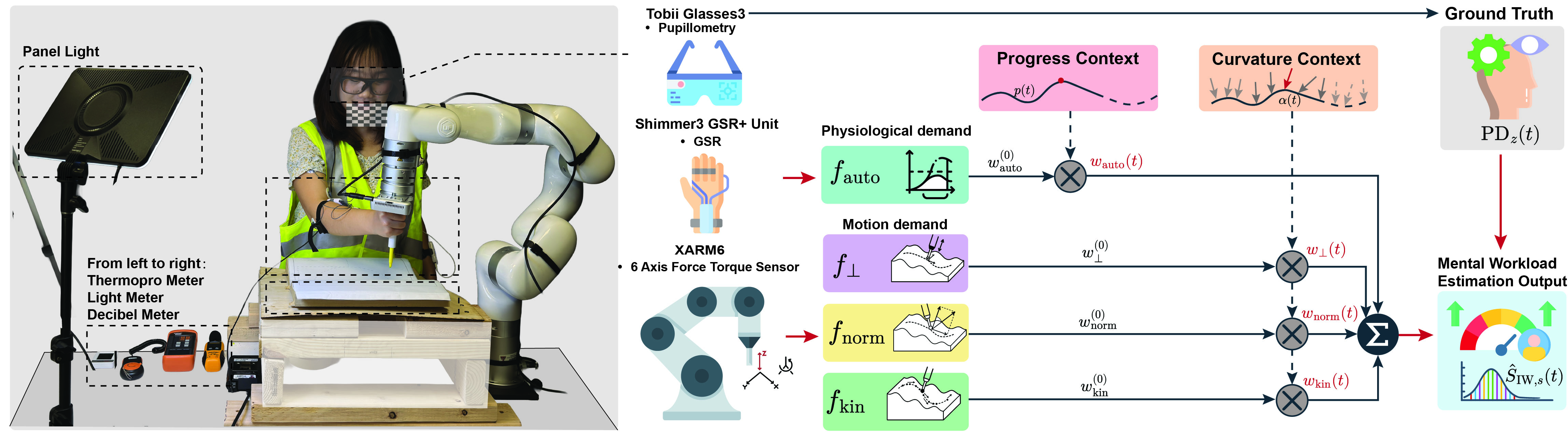}
\caption{Overview of the proposed multimodal interaction workload assessment framework for contact-rich pHRI.}
\label{fig:setup}
\end{figure*}

Fig.~\ref{fig:setup} illustrates the proposed multimodal interaction workload assessment framework. The raw interaction wrenches, TCP kinematics, and GSR  data are transformed into four interpretable workload-related factors. The framework derives three motion-related factors that capture corrective interaction, contact regulation, and kinematic irregularity, while a fourth factor represents tonic autonomic arousal. All four factors are normalized to the $[0,1]$ range and combined using subject-independent base weights. During task execution, the weights of the motion-related factors increase with path curvature context, while the autonomic-arousal weight decreases with task progression context. For details, let $\overline{(\cdot)}_{\Delta T}$ and $\sigma_{\Delta T}(\cdot)$ denote the causal rolling mean and standard deviation over a window of duration $\Delta T$. Motion-derived factors use a shorter window $\Delta T_m$ to reflect short-term interaction changes, whereas SCL uses a longer window $\Delta T_a$ to capture slower tonic variation.

\subsection{\textbf{Workload-related Factors Design}}

In contact-rich pHRI, the operator continuously responds to robot motion and interaction forces while regulating task execution. The motion-derived factors characterize workload-related variation in physical interaction behavior, while the autonomic-arousal factor provides complementary physiological information. Their combination is used to represent interaction workload during physically coupled task execution. Let $\mathbf{F}=(F_x,F_y,F_z)$ denote the interaction force, with planar component $\mathbf{F}_{xy}=(F_x,F_y)$, and let $\mathbf{v}_{xy}=(v_x,v_y)$ denote the planar TCP velocity. The local surface normal is denoted by $\hat{\mathbf{z}}$.

\textbf{Cross-track correction factor ($f_{\perp}$)} captures lateral corrective interaction during path following. Defining the direction perpendicular to the TCP motion as $\mathbf{u}_{\perp}=(-v_y,v_x)/|\mathbf{v}_{xy}|$,
\begin{equation}
F_{\perp}(t)=\mathbf{F}_{xy}(t)\cdot\mathbf{u}_{\perp}(t),
\qquad
f_{\perp}(t)=\sigma_{\Delta T_m}\!\left(F_{\perp}(t)\right).
\label{eq:iw_f_perp}
\end{equation}

The perpendicular force isolates the lateral component of the interaction force from the component along the TCP motion direction. Greater local variation indicates larger fluctuations in lateral force during path following, reflecting greater lateral corrective effort required by the operator to maintain tracking accuracy.

\textbf{Contact-regulation factor ($f_{\mathrm{norm}}$)} measures how much of the interaction force is not aligned with the surface normal. With $F_n=\mathbf{F}\cdot\hat{\mathbf{z}}$,
\begin{equation}
f_{\mathrm{norm}}(t)
=
1-
\overline{
\dfrac{|F_n(t)|}{|\mathbf{F}(t)|}
}_{\Delta T_m}.
\label{eq:iw_f_norm}
\end{equation}
A larger value indicates a greater tangential-force contribution, reflecting greater operator effort to correct the interaction force and maintain consistent surface contact.

\textbf{Kinematic irregularity factor ($f_{\mathrm{kin}}$)} measures abrupt changes in planar TCP motion using translational jerk, $\mathbf{j}_{xy}=\mathrm{d}^{2}\mathbf{v}_{xy}/\mathrm{d}t^{2}$:
\begin{equation}
f_{\mathrm{kin}}(t)
=
\overline{|\mathbf{j}_{xy}(t)|}_{\Delta T_m}.
\label{eq:iw_f_kin}
\end{equation}

A larger value indicates less fluent and more irregular TCP motion during task execution. Prior studies have reported that reduced motion smoothness in human–robot interaction (HRI) is associated with increased physiological stress~\cite{rojas2019}. Accordingly, a higher $f_{\mathrm{kin}}$ is interpreted as indicating greater operator workload.

\textbf{Autonomic-arousal factor ($f_{\mathrm{auto}}$)} characterizes tonic autonomic arousal during task execution using EDA. Raw GSR is low-pass filtered at \SI{2}{Hz} following the preprocessing pipeline in~\cite{DENG2021108098}, and Continuous Decomposition Analysis(CDA) is applied to separate tonic and phasic activity. Only tonic SCL is retained. EDA reflects sympathetic activation but is also affected by individual differences and environmental conditions, particularly temperature. To reduce these effects, each experimental run uses its own resting calibration interval collected beforehand under the same environmental condition. With $\mu_{\mathrm{SCL}}$ denoting the mean SCL during the corresponding calibration interval,
\begin{equation}
f_{\mathrm{auto}}(t)
=
\max\!\left(
0,\,
\overline{\mathrm{SCL}}_{\Delta T_a}(t)
-
\mu_{\mathrm{SCL}}
\right).
\label{eq:iw_f_auto}
\end{equation}
A larger value indicates a greater increase in tonic SCL relative to the local baseline and therefore higher autonomic arousal during interaction. This increase serves as a physiological workload-related indicator.

\subsection{\textbf{Normalization}}

Because the four workload-related factors differ in physical units and numerical ranges, each factor is normalized using a reference scale derived from an independent pilot dataset. The same factor computations described above are applied to the pilot task data to obtain a distribution $\mathcal{F}_{i}^{\mathrm{pilot}}$ for each factor $i\in\{\perp,\mathrm{norm},\mathrm{kin},\mathrm{auto}\}$.

The normalization scale for each factor is defined as
\begin{equation}
\tau_i
=
Q_{0.90}\!\left(\mathcal{F}_{i}^{\mathrm{pilot}}\right),
\qquad
i\in\{\perp,\mathrm{norm},\mathrm{kin},\mathrm{auto}\},
\label{eq:iw_tau}
\end{equation}
where the 90th percentile is used to capture representative peak values while remaining robust against transient outliers. The resulting scales are fixed before the main evaluation and applied to all participants. The normalized factors are
\begin{equation}
\tilde{f}_{i,s}(t)
=
\left[
\dfrac{f_{i,s}(t)}{\tau_i}
\right]_{0}^{1},
\qquad
[x]_0^1:=\min\{1,\max\{0,x\}\}.
\label{eq:iw_tilde_f}
\end{equation}
Values above the pilot-derived reference scale are clipped at one, providing a common $[0,1]$ range for subsequent multimodal fusion.

\subsection{\textbf{Base-Weight Estimation via Ridge Regression}}

The relative contribution of the four workload-related factors is estimated using ridge regression under a strict LOSO scheme. For each evaluation participant $s$, the regression model is fitted using data from all remaining participants, with the four normalized factors as predictors and the physiological reference $\mathrm{PD}_z(t)$ as the target. Data from participant $s$ are excluded from weight estimation. Within each LOSO fold, the predictors are standardized using statistics computed from the training participants only. Ridge regression is then fitted with regularization coefficient $\lambda_r$. The resulting coefficients are mapped back to the normalized-factor scale, negative coefficients are set to zero, and the weight vector is normalized subject to a minimum factor weight $w_{\min}$. The resulting base weights $w_{i,s}^{(0)}$ are fixed before evaluating participant $s$.

\subsection{\textbf{Contextual Adaptation}}

During task execution, path geometry and task progression can change the relative relevance of individual workload-related factors, making fixed fusion weights less suitable across the entire task. To account for this context dependence, an online adaptor adjusts the offline-estimated LOSO ridge-regression base weights $w_{i,s}^{(0)}$ using real-time contextual signals. The geometric context is defined as
\begin{equation}
\alpha_s(t)
=
\left[
\dfrac{\bar{\kappa}(t)}
{\kappa_{90}^{(-s)}}
\right]_0^1,
\label{eq:iw_alpha}
\end{equation}
where $\bar{\kappa}(t)$ is the causal rolling mean of the measured planar TCP curvature over a predefined context window. The reference $\kappa_{90}^{(-s)}$ is the 90th percentile of the corresponding curvature values from the LOSO training participants and is fixed before evaluating participant $s$, ensuring that neither future data nor held-out participant data contribute to the curvature scale. Higher-curvature segments require greater directional adjustment during path following and therefore increase the relevance of the motion-derived factors~\cite{viviani1995minimum}.

Task progression is represented by
\begin{equation}
p(t)=\min\!\left(1,\frac{t}{T_{\mathrm{ref}}}\right),
\label{eq:iw_progress}
\end{equation}
where $T_{\mathrm{ref}}^{(-s)}$ is the median completion time computed from the LOSO training participants and fixed before evaluating participant $s$. Because tonic SCL may drift gradually over time, the progression term downweights $f_{\mathrm{auto}}$ to reduce the influence of slow SCL drift. Overall, the context-adapted weights are
\begin{equation}
\begin{aligned}
w_{i,s}(t)
&=
w_{i,s}^{(0)}
\bigl(1+\lambda_i\alpha_s(t)\bigr),
&& i\in\mathcal{M},\\
w_{\mathrm{auto},s}(t)
&=
w_{\mathrm{auto},s}^{(0)}
\bigl(1-\lambda_p p(t)\bigr).
\end{aligned}
\label{eq:iw_weights}
\end{equation}

The final interaction-workload estimate is computed as
\begin{equation}
\hat{S}_{\mathrm{IW},s}(t)
=
\dfrac{
\sum_i w_{i,s}(t)\tilde{f}_{i,s}(t)
}{
\sum_i w_{i,s}(t)
}.
\label{eq:iw_score}
\end{equation}
Because the normalized factors are bounded and the weights remain nonnegative, $\hat{S}_{\mathrm{IW},s}(t)\in[0,1]$. Normalization by the total weight ensures that contextual adaptation changes the relative contribution of each factor rather than mechanically increasing the interaction-workload score when multiple weights increase simultaneously. Remaining implementation parameters are reported in Section~\ref{sec:para}.
\section{Experiments}
\label{sec:Experiments}

\subsection{Experiment Protocol}
\label{sec:methods_design}

\begin{figure*}[!t]
  \centering
  \includegraphics[width=\textwidth]{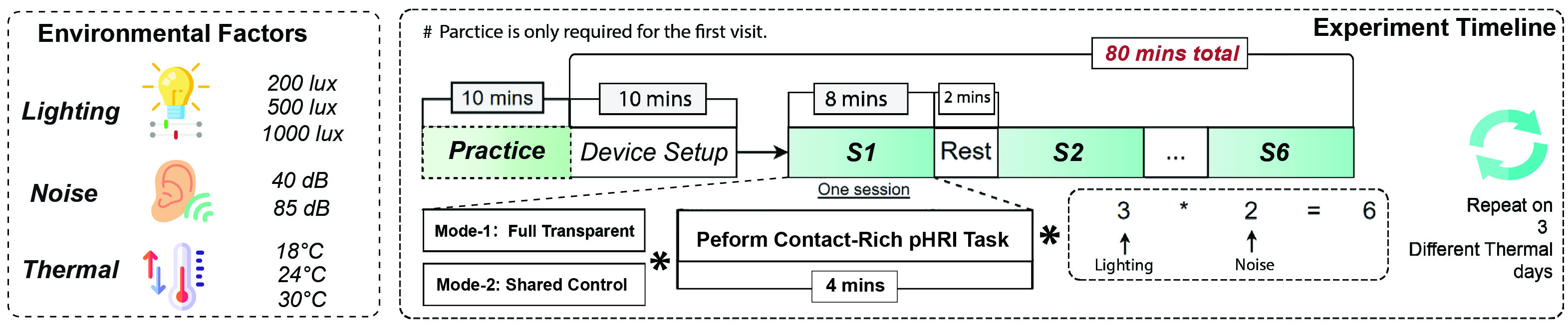}
  \caption{Experimental protocol overview.}
  \label{fig:protocol}
\end{figure*}

\begin{figure}[!b]
  \centering
  \includegraphics[width=\linewidth]{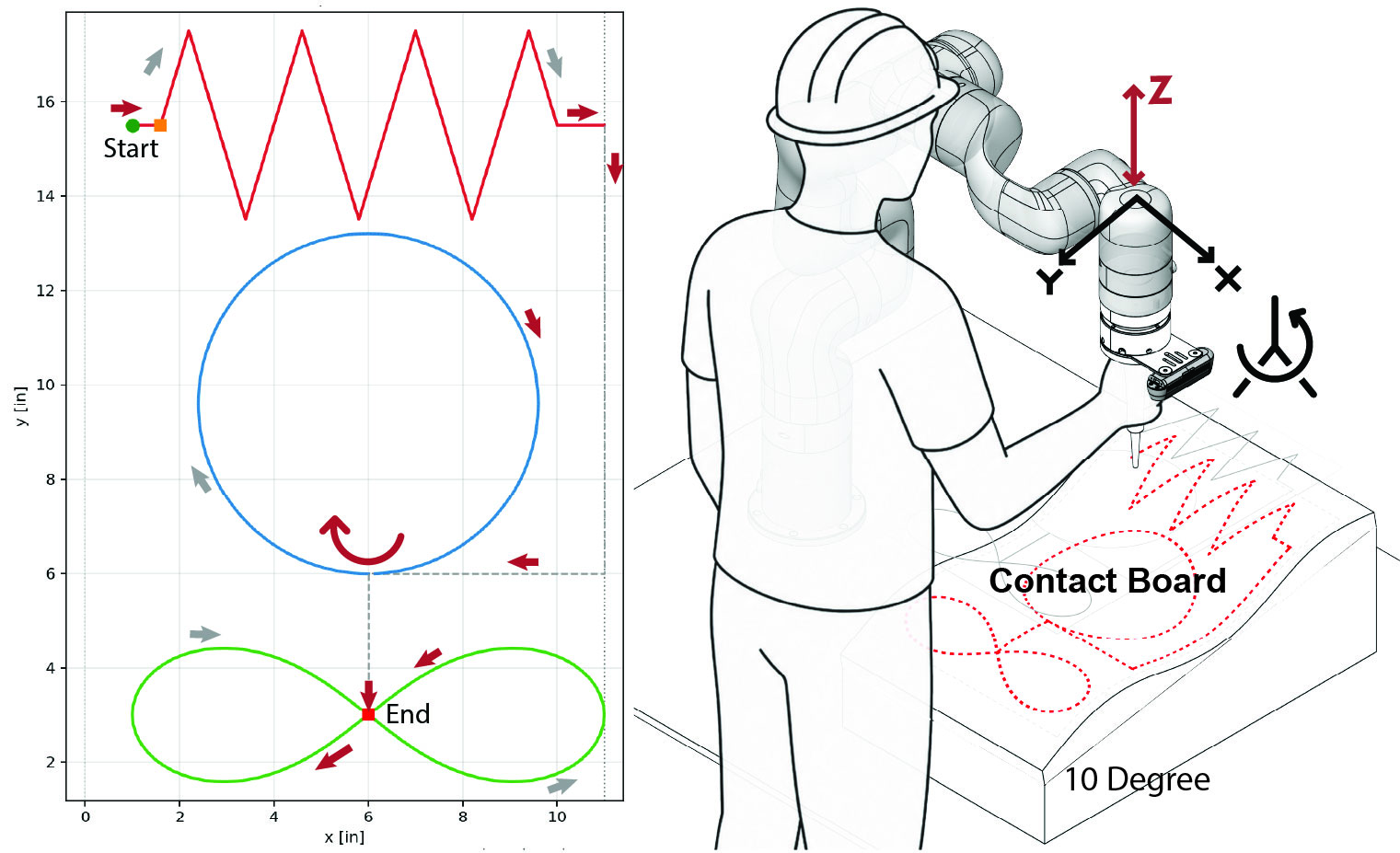}
  \caption{Contact-rich tracing task: nominal four-segment path, motion direction, and an example scene on the inclined corrugated surface.}
  \label{fig:task_traj}
\end{figure}

Figure~\ref{fig:protocol} summarizes the overall experimental procedure and the corresponding full-factorial condition matrix. A $3\times2\times3\times2$ ($Thermal\times Noise\times Lighting\times mode$) within-subject design was adopted to examine participant responses across a range of controlled environmental conditions and two different robot control modes. Data collection was distributed over three laboratory visits conducted on separate days. Each visit was assigned one thermal condition (either 18, 24, or \SI{30}{\celsius}), representing cool, neutral, and warm environments. These temperature settings were selected with reference to ASHRAE guidelines, while relative humidity was maintained at approximately 45\% throughout the experiment~\cite{ashrae2017}.

Within each visit, participants completed six environmental sessions covering all combinations of three illuminance levels (\SIlist{200;500;1000}{lux}) and two acoustic noise levels (\SIlist{40;85}{dB}). The lighting system was maintained at a color temperature of 5000~K. The illuminance levels were selected based on prior human-factors research to represent lighting conditions associated with basic physical activities, standard office work, and precision drawing tasks~\cite{DENG2021108098}. The two acoustic conditions represented a relatively quiet baseline and a high-noise industrial environment, with the upper level selected in accordance with occupational noise exposure guidance from NIOSH~\cite{chan1998occupational}.

Before data collection, participants completed a \SI{10}{min} practice during the first visit to become familiar to interacting with the robotic system. Each visit began with approximately \SI{15}{min} of device setup and signal-quality verification, followed by the six environmental sessions. Under each environmental condition, participants completed two \SI{4}{min} contact-rich tracing trials, one under each robot control mode. Across the three visits, each participant therefore completed 36 tracing trials in total.

\subsection{Participants}
\label{sec:participants}

A total of 27 healthy adults participated in the study ($N=27$; mean age $28.8 \pm 4.5$ years; 12 female and 15 male). Participants were recruited from the local community and consisted primarily of graduate students. Eligibility required participants to be capable of safely performing light physical activities. Individuals were excluded if they reported musculoskeletal, neurological, or cardiovascular conditions. Participants with skin sensitivities at the intended GSR electrode placement were also excluded. The experimental protocol was reviewed and approved by the Institutional Review Boards (IRBs) of Texas Tech University and the University of Tennessee, Knoxville. All participants provided written informed consent before data collection and were informed of their right to withdraw from the study at any time. Following data-quality screening, complete datasets from 24 participants were retained for subsequent analysis ($N=24$; mean age $28.4 \pm 4.4$ years; 11 female and 13 male). 

\subsection{Experiment Setup}

\subsubsection{\textbf{Task Setup}}
\label{sec:task}

The task scenario is illustrated in Fig.~\ref{fig:task_traj}. Participants grasped a custom end-effector and guided a marker along a predefined path on a contact board while maintaining continuous marker-surface contact. The board had a mild sinusoidal corrugation along the primary movement direction and was inclined by approximately \SI{10}{\degree} from the horizontal plane. Participants were instructed to trace the full path as accurately as possible. The path consisted of a zigzag, a long straight line, a circle, and a Bernoulli lemniscate, providing sharp directional changes, steady linear motion, continuous curvature, and self-crossing geometry.

\subsubsection{\textbf{Apparatus and Environment Setup}}
\label{sec:platform}

The collaborative task was performed using a 6-DOF UFactory xArm6 equipped with a wrist-mounted 6-axis force-torque sensor for interaction-wrench measurement. EDA was recorded using a wireless Shimmer3 GSR+ unit worn on the left arm, while Tobii Pro Glasses~3 recorded pupillometry for offline validation. The robot was operated at a safe kinematic scale, and an emergency stop was accessible to both the participant and the experimenter.

All experiments were conducted in a sealed laboratory with the central HVAC system disabled. Temperature was controlled using portable heating and cooling devices. Participants wore long trousers and short-sleeved shirts, corresponding to approximately \SI{0.5}{clo} based on ASHRAE thermal-comfort guidelines. Illuminance was adjusted using an LED panel, with blackout curtains used to block daylight. The baseline acoustic level was approximately \SI{40}{dB}, while the high-noise condition was generated by playing industrial white noise at \SI{85}{dB}. Environmental conditions were continuously monitored using calibrated meters: temperature at abdomen height ($\pm\SI{0.5}{\celsius}$), acoustic level near ear height ($\pm\SI{5}{dB}$), and illuminance at the center of the workspace ($\pm\SI{30}{lux}$).

\subsubsection{\textbf{Robot Setup}}
\label{sec:control}

To evaluate our framework across distinct physical interaction conditions, the robot was operated under two admittance-control modes: \textbf{Full Transparency (Mode 1)} and \textbf{Orthogonal Shared Control (Mode 2)}. In both modes, the robot operates in Cartesian velocity mode under a zero-stiffness ($k_v=0$) admittance controller:
\begin{equation}
\mathbf{M}_v \dot{\mathbf{v}}_d
+
\mathbf{B}_v \mathbf{v}_d
=
\mathbf{u},
\label{eq:virtual_dynamics}
\end{equation}
where $\mathbf{v}_d\in\mathbb{R}^6$ is the velocity command, and $\mathbf{M}_v$ and $\mathbf{B}_v$ are diagonal virtual inertia and damping matrices configured as $\mathbf{M}_v=\operatorname{diag}(1,1,1,0.15,0.15,0.15)$ and $\mathbf{B}_v=\operatorname{diag}(120,120,450,4.5,4.5,4.5)$. To suppress oscillations during direction changes, the effective damping $B_{v,i}$ is scaled by a reversal damping factor $\beta=\num{1.8}$ whenever the commanded input opposes the current velocity along axis $i$ ($u_i v_{d,i}<0$). The admittance input $\mathbf{u}=[\mathbf{u}_{xy}^{\top},u_z,\mathbf{u}_{\tau}^{\top}]^{\top}\in\mathbb{R}^6$ is synthesized from the measured external wrench $\mathbf{W}_{\mathrm{ext}}=[\mathbf{F}_{\mathrm{ext}}^{\top},\boldsymbol{\tau}_{\mathrm{ext}}^{\top}]^{\top}$. Control parameters and velocity limits were empirically tuned through pilot testing.

\subsubsection{\textbf{Mode 1: Full Transparency}}

In this mode, the robot acts as a transparent manipulator across all spatial degrees of freedom. To suppress sensor noise and unintended tremor, a deadband scaling function $\phi(f)\in[0,1]$ is applied to the magnitude of the measured translational force to suppress small force inputs. For $f\leq F_{\mathrm{dz}}$, the input is suppressed and $\phi(f)=0$. Over the transition range $F_{\mathrm{dz}}<f<F_{\mathrm{dz}}^{U}$, the scaling factor increases from $0$ to $1$. For $f\geq F_{\mathrm{dz}}^{U}$, $\phi(f)=1$, the measured force passes through without attenuation. The lower and upper thresholds are set to $F_{\mathrm{dz}}=\SI{2.8}{N}$ and $F_{\mathrm{dz}}^{U}=\SI{3.6}{N}$. The planar admittance input is $\mathbf{u}_{xy}=G_{xy}\phi(|\mathbf{F}_{\mathrm{ext},xy}|)\mathbf{F}_{\mathrm{ext},xy}$, and the normal input is $u_z=G_z\phi(|F_{\mathrm{ext},z}|)F_{\mathrm{ext},z}$, with $G_{xy}=G_z=\num{0.55}$.

In the rotational subspace, the input is set to zero when the magnitude of the external torque $\tau=|\boldsymbol{\tau}_{\mathrm{ext}}|$ is below the deadband threshold $\tau_{\mathrm{th}}$ and to $G_{\tau}\boldsymbol{\tau}_{\mathrm{ext}}$ otherwise, where $G_{\tau}=\num{0.28}$ and $\tau_{\mathrm{th}}=\SI{0.1}{N\,m}$. Translational velocities are limited to $\pm v_{\max}=\SI{38}{mm/s}$, and rotational velocities are limited to $\pm\omega_{\max}=\SI{22}{deg/s}$.The deadband thresholds, admittance gains, and velocity limits were selected empirically through pilot testing before the main experiment.

\textbf{In Mode 2}, the task space is orthogonally decoupled into a human-guided subspace and a robot-regulated normal subspace ($Z$). The participant guides the planar trajectory, while the robot regulates the normal contact force at $F_d=\SI{5}{N}$ against the unknown corrugated surface. Mode 2 uses the same controller configuration as the orthogonal shared-control architecture reported in our previous work~\cite{chen2026multimodal}.

\subsubsection{\textbf{Implementation Details}}
\label{sec:para}

The short causal motion window was set to $\Delta T_m=3\,\mathrm{s}$, and the longer autonomic window was set to $\Delta T_a=15\,\mathrm{s}$. Each experimental run used a $15\,\mathrm{s}$ resting interval for the local SCL baseline. The contextual gains were set to $\lambda_{\mathrm{kin}}=1.0$, $\lambda_{\perp}=\lambda_{\mathrm{norm}}=0.2$, and $\lambda_p=0.6$. Across the 24 LOSO folds, the mean learned base weights were $\bar{w}{\perp}^{(0)}=0.394$, $\bar{w}{\mathrm{norm}}^{(0)}=0.227$, $\bar{w}{\mathrm{kin}}^{(0)}=0.213$, and $\bar{w}{\mathrm{auto}}^{(0)}=0.166$. 

\subsection{Offline Validation and Evaluation Protocol}
\label{sec:metric}

\subsubsection{\textbf{Physiological Reference}}

Pupillometry was used as an independent physiological reference for offline validation~\cite{aygunetal22sensors}. Binocular pupil diameter was recorded using Tobii Pro Glasses~3 at $100\,\mathrm{Hz}$. Samples without valid two-dimensional gaze data were discarded, and missing pupil measurements were represented as NaN. At each timestamp, available left- and right-eye measurements were averaged, retaining a single valid eye when the other was unavailable. Duplicate timestamps were merged using a NaN-aware mean, and the resulting pupil series was interpolated to $50\,\mathrm{Hz}$.

For each experimental run, the preceding calibration interval was used to compute the baseline pupil mean $d_0$ and standard deviation $\sigma_0$. The calibration interval and subsequent tracing task were conducted under the same environmental conditions.  A causal moving average was then evaluated at $1\,\mathrm{Hz}$ to align the pupil signal with our framework. The standardized pupil diameter is defined as
\begin{equation}
\mathrm{PD}_z(t)
=
\frac{d(t)-d_0}{\sigma_0}.
\label{eq:pupil_reference}
\end{equation}
$\mathrm{PD}_z$ is used only as an offline physiological reference.

\subsubsection{\textbf{Comparative Baselines}}

Two groups of comparisons were evaluated under the same strict LOSO protocol. The first group examines the contribution of contextual adaptation within the proposed framework. All four variants use the same normalized workload-related factors and the same fold-specific LOSO base weights, differing only in the contextual reweighting applied during inference. \textit{Static} disables both contextual adaptors, \textit{Static + Curvature} applies only curvature-based modulation, \textit{Static + Progress} applies only progression-based modulation.

The second group consists of three representative state-of-the-art learning-based baselines: MLP, TCN, and LSTM. All models use the same four factors and identical LOSO training and test partitions. The MLP predicts $\mathrm{PD}_z$ from the factors at the current time step, whereas the TCN and LSTM use the preceding $30\,\mathrm{s}$ of factor history. To harmonize their unbounded outputs with the $[0,1]$ range of the proposed estimator, the network outputs are rank-mapped to $[0,1]$ within each held-out trial for offline evaluation. 
\begin{table*}[!t]
\centering
\caption{Comparison of Framework Variants and Learning-Based Baselines}
\label{tab:comparison_results}

\begin{threeparttable}
\small
\resizebox{0.99\textwidth}{!}{%
\begin{tabular}{@{}llccccccc@{}}
\toprule
\footnotesize
Category
& Method
& Pos.\,Subj.
& Pos.\,Trial
& $\rho_{30}$
& Median $\rho_{30}$ [IQR]
& $\rho_{\mathrm{early}}$
& $\rho_{\mathrm{mid}}$
& $\rho_{\mathrm{late}}$ \\
\midrule

\multirow{4}{*}{Adaptor variant}
& Static
& 23/24 (95.8\%)
& 73.0\%
& 0.269*
& 0.292 [0.138, 0.359]
& 0.023
& 0.092
& 0.062 \\

& Static + Curvature
& 23/24 (95.8\%)
& 73.5\%
& 0.288
& 0.296 [0.160, 0.349]
& 0.007
& 0.099
& 0.067 \\

& Static + Progress
& 23/24 (95.8\%)
& 75.2\%
& 0.298
& 0.297 [0.167, 0.402]
& 0.017
& 0.101
& 0.073 \\

& \textbf{Ours}
& \textbf{23/24 (95.8\%)}
& \textbf{75.4\%}
& \textbf{0.308}
& \textbf{0.314 [0.188, 0.405]}
& 0.002
& 0.112
& 0.068 \\

\midrule

\multirow{3}{*}{Learning baseline}
& MLP
& \textbf{23/24 (95.8\%)}
& \textbf{75.9\%}
& \textbf{0.313}
& \textbf{0.341 [0.195, 0.372]}
& 0.015
& 0.080
& 0.001 \\

& TCN
& 16/24 (66.7\%)
& 56.4\%
& 0.058**
& 0.110 [$-0.122$, 0.237]
& $-0.009$
& 0.034
& 0.042 \\

& LSTM
& 19/24 (79.2\%)
& 61.7\%
& 0.175*
& 0.156 [0.040, 0.341]
& 0.014
& 0.013
& 0.089 \\

\bottomrule
\end{tabular}%
}

\begin{tablenotes}[flushleft]
\footnotesize
\item *$p_{\mathrm{Holm}}<0.05$, **$p_{\mathrm{Holm}}<0.01$ for two-sided paired Wilcoxon signed-rank comparisons with Ours based on subject-level $\rho_{30,s}$. Bold values indicate the best result within each method category.
\end{tablenotes}

\end{threeparttable}
\end{table*}
\subsubsection{\textbf{Evaluation Metrics}}

Performance was evaluated by the correspondence between the estimated interaction workload and the physiological reference at the trial, subject, and task-progress levels. The primary metric was the $30\,\mathrm{s}$ block-wise Spearman correlation($\rho_{30,s}$). For each valid trial, the workload output and $\mathrm{PD}_z$ were averaged within non-overlapping $30\,\mathrm{s}$ windows, and Spearman correlation was computed between the resulting block means. The subject-level value $\rho_{30,s}$ was then obtained by averaging the trial-level correlations for participant $s$, and the cohort-level $\rho_{30}$ was calculated as the mean of $\rho_{30,s}$ across participants. The median and interquartile range (IQR) of the subject-level $\rho_{30,s}$ values are also reported.

Cross-subject consistency is summarized by \textit{Pos.Subj.}, defined as the proportion of participants with $\rho_{30,s}>0$. Trial-level consistency is summarized by \textit{Pos.Trial}, defined as the proportion of valid trials with positive $30\,\mathrm{s}$ block-wise Spearman correlation. These measures characterize how consistently positive correspondence is observed across participants and individual trials.

To examine whether the correspondence remains consistent throughout the task rather than being concentrated in a specific phase, each trial was divided into early, middle, and late stage of normalized progress, corresponding to $[0,33\%)$, $[33,67\%)$, and $[67,100\%)$. The corresponding correlations are denoted by $\rho_{\mathrm{early}}$, $\rho_{\mathrm{mid}}$, and $\rho_{\mathrm{late}}$.

Pairwise method comparisons were performed at the subject level using two-sided paired Wilcoxon signed-rank tests on $\rho_{30,s}$ across the proposed framework, its contextual variants, and the learning-based baselines.

\section{Experimental Results}
\label{sec:results}

\begin{figure}[!t]
\centering
\includegraphics[width=\linewidth]{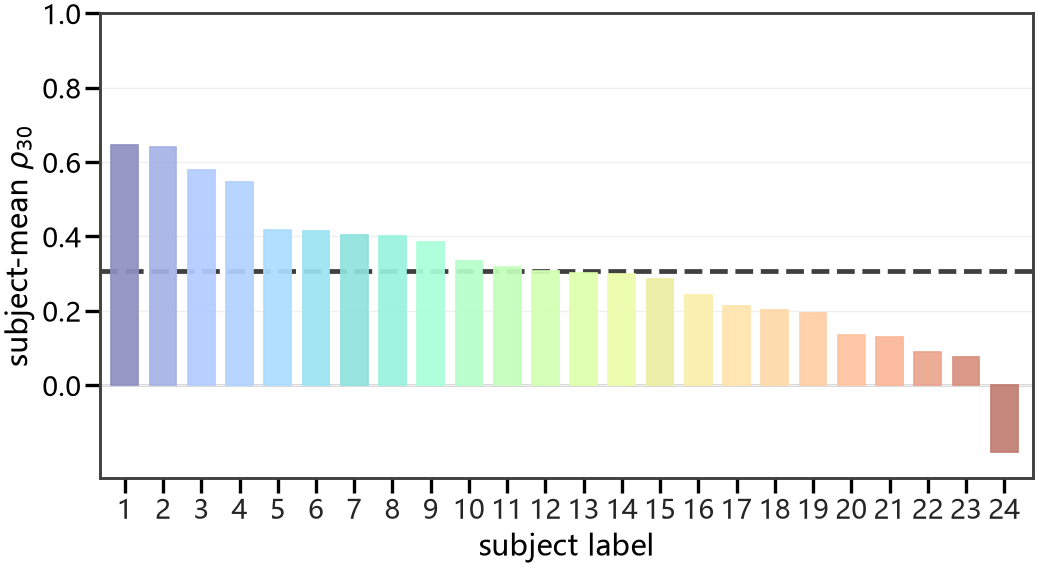}
\caption{
Subject-level $30\,\mathrm{s}$ block-wise Spearman correlation $\rho_{30}$ of our framework, sorted in descending order.
}
\label{fig:subject_rho30_grouped}
\end{figure}

\subsection{Performance of the Proposed Framework}
\label{sec:results_ours}

Table~\ref{tab:comparison_results} summarizes the performance of the proposed framework, its contextual-ablation variants, and the learning-based baselines. Under strict LOSO evaluation, the proposed framework achieves a cohort-mean $30\,\mathrm{s}$ block-wise correlation of $\rho_{30}=0.308$ with the physiological reference, indicating a consistent positive correspondence at the cohort level. Positive subject-level correspondence is observed for 23 of 24 participants (95.8\%), and 75.4\% of valid trials exhibit positive trial-level correspondence. Figure~\ref{fig:subject_rho30_grouped} shows the correlation achieved by the proposed framework for each participant, with positive subject-level values ranging from $0.077$ to $0.644$. The subject-level median correlation is $0.314$ [IQR: $0.188$, $0.405$], showing that the positive correspondence is broadly distributed across participants despite inter-individual variation.

\begin{figure}[!t]
\centering
\includegraphics[width=\linewidth]{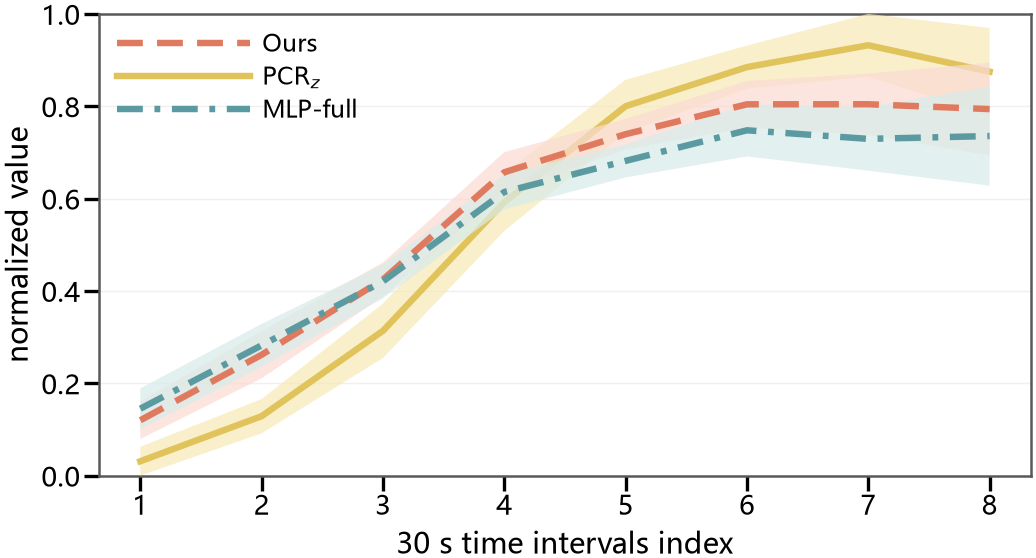}
\caption{
Cohort-mean within-trial $30\,\mathrm{s}$ block for the proposed framework, MLP baseline, and physiological reference $\mathrm{PD}_z$ ($n=24$).}
\label{fig:progress_block30}
\end{figure}

When the task is divided into early, middle, and late portions, the corresponding correlations are $0.002$, $0.112$, and $0.068$. Correspondence is weak near task onset, increases during the middle portion of the task, and remains positive in the late phase. Fig.~\ref{fig:progress_block30} shows that the proposed framework captures several of the same within-trial directional changes as the physiological reference, particularly in the middle portion of the task.

Performance was also examined across the two tested interaction modes. Mode~1 yields $\rho_{30}=0.310$, with 21/24 participants (87.5\%) showing positive subject-level correspondence and 77.0\% of trials showing positive trial-level correspondence. Mode~2 yields $\rho_{30}=0.315$, with 23/24 participants (95.8\%) and 73.1\% of trials showing positive correspondence. Paired Wilcoxon tests detect no significant difference between the two modes in subject-level $\rho_{30}$ ($p=0.41$) or positive-trial rate ($p=0.23$).

\begin{figure}[!t]
\centering
\includegraphics[width=\linewidth]{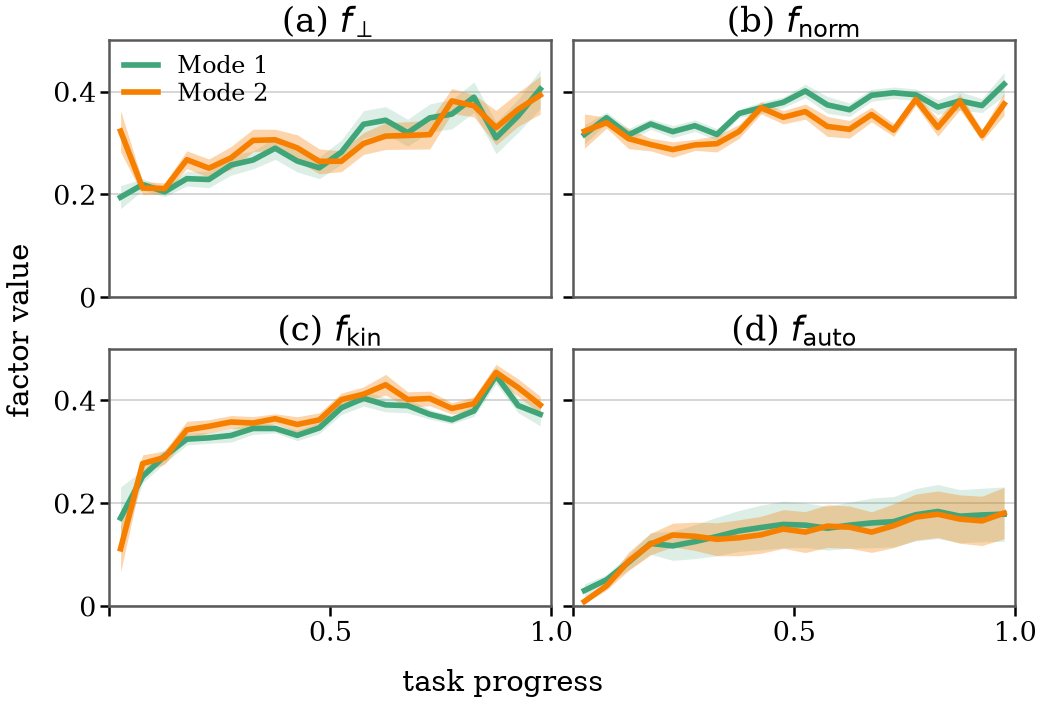}
\caption{
Mean $\pm$ SEM profiles of
(a)~$f_{\perp}$,
(b)~$f_{\mathrm{norm}}$,
(c)~$f_{\mathrm{kin}}$, and
(d)~$f_{\mathrm{auto}}$
under Mode~1 and Mode~2.
}
\label{fig:factor_profiles_4row}
\end{figure}

A similar comparison was conducted for the four normalized workload-related factors. Figure~\ref{fig:factor_profiles_4row} shows how the four factors vary with task progression under the two interaction modes. The mean absolute between-mode differences are $0.026$ for $f_{\perp}$, $0.030$ for $f_{\mathrm{norm}}$, $0.021$ for $f_{\mathrm{kin}}$, and $0.009$ for $f_{\mathrm{auto}}$, with corresponding profile Spearman correlations of $0.79$, $0.76$, $0.97$, and $0.94$. Relative to the normalized $[0,1]$ range, these mean absolute differences are small, while the high correlations indicate similar temporal variation across the two interaction modes.

\begin{table}[!t]
\centering
\caption{Performance Across the Two Tested Interaction Modes}
\label{tab:mode_stability}
\begin{threeparttable}
\small
\setlength{\tabcolsep}{3.5pt}
\renewcommand{\arraystretch}{1.1}
\begin{tabularx}{\linewidth}{@{} l
>{\centering\arraybackslash}X
>{\centering\arraybackslash}X
>{\centering\arraybackslash}X @{}}
\toprule
& $\rho_{30}$
& Pos.\,Subj.
& Pos.\,Trial \\
\midrule
Mode 1
& 0.310
& 21/24 (87.5\%)
& 77.0\% \\
Mode 2
& 0.315
& 23/24 (95.8\%)
& 73.1\% \\
\bottomrule
\end{tabularx}
\end{threeparttable}
\end{table}

\subsection{Performance Comparisons of Baseline }
\label{sec:results_baselines}

\subsubsection{\textbf{Ablation of Contextual Adaptation}}
The four settings in Table~\ref{tab:comparison_results} use the same normalized workload-related factors and LOSO-trained base weights, differing only in the contextual reweighting applied during inference. All four settings show positive subject-level correspondence for 23 of 24 participants (95.8\%). The positive-trial rate increases from 73.0\% for Static to 73.5\% for Static + Curvature, 75.2\% for Static + Progress, and 75.4\% for our model. The cohort-mean $\rho_{30}$ also increases from $0.269$ for Static to $0.288$, $0.298$, and $0.308$, respectively.

The subject-level difference between our model and Static remains significant after Holm correction ($p_{\mathrm{Holm}}=0.036$). In comparison, the differences between our model and Static + Curvature and between our model and Static + Progress are not significant ($p_{\mathrm{Holm}}=0.28$ for both). These results show that adding either curvature or progress information increases the cohort-mean correlation relative to Static, while using both gives the highest $\rho_{30}$. The significant difference between our model and Static supports the benefit of contextual reweighting over fixed weights.

\subsubsection{\textbf{Learning-Based Baselines}}
Among the learning-based baselines, MLP achieves the highest cohort-mean correlation, with $\rho_{30}=0.313$, compared with $0.308$ for Our model. The subject-level difference between the two methods is not significant after Holm correction ($p_{\mathrm{Holm}}=0.88$). Both methods show positive correspondence for 23 of 24 participants (95.8\%), while the positive-trial rates are 75.9\% for MLP and 75.4\% for our model. These results show that our model achieves performance comparable to MLP under the same LOSO evaluation.

TCN and LSTM yield lower cohort-mean correlations of $0.058$ and $0.175$, with lower positive-subject and positive-trial rates. The subject-level differences relative to our model remain significant after Holm correction for TCN ($p_{\mathrm{Holm}}=0.0021$) and LSTM ($p_{\mathrm{Holm}}=0.025$).

The progress-dependent correlations for MLP are $0.015$, $0.080$, and $0.001$ for the early, middle, and late portions of the task.  As shown in Fig.~\ref{fig:progress_block30}, the physiological reference rises again later in the task, while the MLP output follows this change less closely. Compared with the results for our model reported above, MLP shows similar overall correspondence but weaker correspondence in the middle and late portions of the task. In summary, these results show comparable overall performance to the advanced learning-based baseline while retaining the interpretable structure of the proposed framework.

\section{Discussion}
\label{sec:discussion}

The comparison with MLP shows that the proposed framework achieves overall correspondence comparable to that of MLP under the same LOSO evaluation. The progress-dependent analysis also shows different temporal behavior between the two methods. Ours maintains positive correspondence through the middle and late portions of the task, whereas the MLP correspondence decreases toward the late phase. At this similar level of performance, our framework provides an interpretable representation of interaction workload. The final assessment can be related to corrective interaction, contact regulation, kinematic irregularity, autonomic arousal, and their context-dependent weights. In contrast, MLP produces its output through a learned nonlinear mapping, making individual inputs less explicit. The proposed framework therefore offers similar overall performance while keeping the estimation process transparent and the contribution of each factor interpretable.

The ablation results further support the contextual reweighting mechanism. Compared with the Static formulation, our framework achieves higher subject-level correspondence, a higher positive-trial rate, and a higher cohort-mean correlation. The subject-level difference also remains significant after Holm correction. The observed improvements suggest that contextual reweighting is more suitable than fixed weighting when the relative importance of motion-related and physiology-related factors changes during contact-rich pHRI task execution.

The proposed framework also shows consistent behavior across the two tested admittance-control modes. No significant mode difference is detected in subject-level correspondence or positive-trial rate, and the normalized factor profiles show small mean absolute differences with high temporal correlations between modes. These results indicate that the proposed framework maintains similar performance and factor behavior across different control settings. Overall, our framework combines performance comparable to the advanced learning-based baseline with an interpretable factor structure, contextual adaptive weighting, and consistent behavior across different control modes. From a practical implementation perspective, the robot provides the motion and force signals, while EDA can be obtained from wearable devices, supporting practical real-time use with limited additional sensing.

Several limitations of this study motivate future work. First, although the proposed framework operates causally, the current evaluation is based on offline analysis. Future studies will examine real-time implementation and evaluation. Second, each experimental run uses a fixed $15\,\mathrm{s}$ resting calibration for SCL baseline correction. Adaptive baseline estimation may reduce this initialization requirement and better accommodate differences in autonomic response. Finally, correspondence varies across participants and over task progression, indicating that a single subject-independent formulation may not fully capture differences across individuals or task phases. Future studies will examine adaptation across participants and task phases, as well as evaluation over longer interactions and a wider range of pHRI tasks.
\section{Conclusion}
\label{sec:conclusion}

This paper presented an online multimodal framework for continuous interaction workload assessment in contact-rich pHRI. The framework combines interaction wrench, planar TCP kinematics, and EDA through interpretable workload-related factors and adjusts their contributions using task context. Under strict LOSO evaluation, the proposed framework showed consistent correspondence with the pupillometry-based physiological reference, achieved performance comparable to MLP and higher than the static weight method, TCN, and LSTM, and maintained stability across the two tested admittance control modes.
\section*{Acknowledgment}
The authors would like to acknowledge the ﬁnancial support for this research received from the U.S. National Science Foundation (NSF) CMMI 2531678. Any opinions and ﬁndings in this paper are those of the authors and do not necessarily represent those of the NSF.

\bibliographystyle{IEEEtran}

\bibliography{ref}

@ARTICLE{fu2025,
  author={Fu, Junling and Maimone, Giorgia and Iovene, Elisa and Zhao, Jianzhuang and Redaelli, Alberto and Ferrigno, Giancarlo and De Momi, Elena},
  journal={IEEE Transactions on Robotics}, 
  title={Human-Inspired Active Compliant and Passive Shared Control Framework for Robotic Contact-Rich Tasks in Medical Applications}, 
  year={2025},
  volume={41},
  number={},
  pages={2549-2568},
  doi={10.1109/TRO.2025.3548493}}

@article{storm2022,
author = {Storm, Fabio A. and Chiappini, Mattia and Dei, Carla and Piazza, Caterina and André, Elisabeth and Reißner, Nadine and Brdar, Ingrid and Delle Fave, Antonella and Gebhard, Patrick and Malosio, Matteo and Peña Fernández, Alberto and Štefok, Snježana and Reni, Gianluigi},
title = {Physical and mental well-being of cobot workers: A scoping review using the Software-Hardware-Environment-Liveware-Liveware-Organization model},
journal = {Human Factors and Ergonomics in Manufacturing \& Service Industries},
volume = {32},
number = {5},
pages = {419-435},
doi = {https://doi.org/10.1002/hfm.20952},
year = {2022}
}

@ARTICLE{polish2025,
  author={Widanage, Kithmi N. D. and Xia, Jingkang and Parween, Rizuwana and Godaba, Hareesh and Herzig, Nicolas and Glovnea, Romeo and Huang, Deqing and Li, Yanan},
  journal={IEEE Transactions on Robotics}, 
  title={Nonrepetitive-Path Iterative Learning and Control for Human-Guided Robotic Operations on Unknown Surfaces}, 
  year={2025},
  volume={41},
  number={},
  pages={4922-4940},
  doi={10.1109/TRO.2025.3588453}}

@article{belkaidMutualGazeRobot2021,
  title = {Mutual Gaze with a Robot Affects Human Neural Activity and Delays Decision-Making Processes},
  author = {Belkaid, Marwen and Kompatsiari, Kyveli and De Tommaso, Davide and Zablith, Ingrid and Wykowska, Agnieszka},
  year = 2021,
  month = sep,
  journal = {Science Robotics},
  volume = {6},
  number = {58},
  pages = {eabc5044},
  publisher = {American Association for the Advancement of Science},
  doi = {10.1126/scirobotics.abc5044}
}

@article{DENG2021108098,
  title = {Measurement and Prediction of Work Engagement under Different Indoor Lighting Conditions Using Physiological Sensing},
  author = {Deng, Min and Wang, Xi and Menassa, Carol C.},
  year = 2021,
  journal = {Building and Environment},
  volume = {203},
  pages = {108098},
  issn = {0360-1323},
  doi = {10.1016/j.buildenv.2021.108098}
}

@article{deng2026integrating,
  title={Integrating LLMs and digital twins for adaptive multi-robot task allocation in construction},
  author={Deng, Min and Fu, Bo and Li, Lingyao and Wang, Xi},
  journal={IEEE Transactions on Automation Science and Engineering},
  year={2026},
  publisher={IEEE}
}

@article{chen2026perception,
  title={From Perception to Symbolic Task Planning: Vision-Language Guided Human-Robot Collaborative Structured Assembly},
  author={Chen, Yanyi and Deng, Min},
  journal={arXiv preprint arXiv:2601.00978},
  year={2026}
}

@manual{ashrae2017,
  author       = {{ASHRAE}},
  title        = {Standard 55-2017, Thermal Environmental Conditions for Human Occupancy},
  address      = {Atlanta, USA},
  year         = {2017}
}

@techreport{chan1998occupational,
  author      = {Chan, H. S.},
  title       = {Occupational noise exposure; criteria for a recommended standard},
  institution = {National Institute for Occupational Safety and Health (NIOSH)},
  year        = {1998}
}

@ARTICLE{see2025ultrasound,
  author={Zhetpissov, Yernar and Ma, Xihan and Yang, Kehan and Zhang, Haichong K.},
  journal={IEEE Robotics and Automation Letters}, 
  title={A-SEE2.0: Active-Sensing End-Effector for Robotic Ultrasound Systems with Dense Contact Surface Perception Enabled Probe Orientation Adjustment}, 
  year={2025},
  volume={10},
  number={9},
  pages={9557-9564},
  doi={10.1109/LRA.2025.3595036}}

@Article{force2024ultrasound,
AUTHOR = {Jiang, Jinlei and Luo, Jingjing and Wang, Hongbo and Tang, Xiuhong and Nian, Fan and Qi, Lizhe},
TITLE = {Force Tracking Control Method for Robotic Ultrasound Scanning System under Soft Uncertain Environment},
JOURNAL = {Actuators},
VOLUME = {13},
YEAR = {2024},
NUMBER = {2},
ARTICLE-NUMBER = {62},
ISSN = {2076-0825},
DOI = {10.3390/act13020062}
}

@ARTICLE{co-carrying2024framework,
  author={Dang, Van Trong and Kotake, Hiroki and Honji, Sumitaka and Wada, Takahiro},
  journal={IEEE Transactions on Automation Science and Engineering}, 
  title={A Cooperation Control Framework Based on Admittance Control and Time-Varying Passive Velocity Field Control for Human–Robot Co-Carrying Tasks}, 
  year={2025},
  volume={22},
  number={},
  pages={23579-23593},
  doi={10.1109/TASE.2025.3628318}}

@article{upasani2023eye,
    author = {Satyajit Upasani and Divya Srinivasan and Qi Zhu and Jing Du and Alexander Leonessa},
    title ={Eye-Tracking in Physical Human–Robot Interaction: Mental Workload and Performance Prediction},
    journal = {Human Factors},
    volume = {66},
    number = {8},
    pages = {2104-2119},
    year = {2024},
    doi = {10.1177/00187208231204704},
    note ={PMID: 37793896},
}

@ARTICLE{progress2023review,
  author={Xue, Teng and Wang, Weiming and Ma, Jin and Liu, Wenhai and Pan, Zhenyu and Han, Mingshuo},
  journal={IEEE Sensors Journal}, 
  title={Progress and Prospects of Multimodal Fusion Methods in Physical Human–Robot Interaction: A Review}, 
  year={2020},
  volume={20},
  number={18},
  pages={10355-10370},
  doi={10.1109/JSEN.2020.2995271}}

@inproceedings{kiki2025estimating,
  title={Estimating Human Muscular Fatigue in Dynamic Collaborative Robotic Tasks with Learning-Based Models},
  author={Kiki, Feras and Niaz, Pouya P. and Madani, Alireza and Basdogan, Cagatay},
  booktitle={arXiv preprint arXiv:2602.15684},
  year={2026}
}

@article{xia2025large,
  title = {A large language model-driven framework for multimodal cognitive workload prediction in industrial human–robot collaboration},
  volume = {181},
  ISSN = {0952-1976},
  DOI = {10.1016/j.engappai.2026.115694},
  journal = {Engineering Applications of Artificial Intelligence},
  publisher = {Elsevier BV},
  author = {Xue,  Qiwei and Zhang,  Yuchong and Wu,  Huapeng and Song,  Yuntao},
  year = {2026},
  month = Oct,
  pages = {115694}
}

@article{peternel2018robot,
  author    = {Peternel, L. and Tsagarakis, N. and Caldwell, D. and others},
  title     = {Robot adaptation to human physical fatigue in human--robot co-manipulation},
  journal   = {Autonomous Robots},
  volume    = {42},
  pages     = {1011--1021},
  year      = {2018},
  doi       = {10.1007/s10514-017-9678-1},
  publisher = {Springer}
}

@article{lorenzini2022ergonomic,
  title = {Ergonomic human-robot collaboration in industry: A review},
  volume = {9},
  ISSN = {2296-9144},
  DOI = {10.3389/frobt.2022.813907},
  journal = {Frontiers in Robotics and AI},
  publisher = {Frontiers Media SA},
  author = {Lorenzini,  Marta and Lagomarsino,  Marta and Fortini,  Luca and Gholami,  Soheil and Ajoudani,  Arash},
  year = {2023},
  month = Jan 
}

@article{aygunetal22sensors,
  title={Investigating Methods for Cognitive Workload Estimation for Assistive Robots},
  author={Aygun, Ayca and Nguyen, Thuan and Haga, Zachary and Aeron, Shuchin and Scheutz, Matthias},
  year={2022},
  journal={Sensors},
  publisher={MDPI},
  volume={22},
  pages={6834}
}

@ARTICLE{cakit2025,
  author={Çakit, Erman and Karwowski, Waldemar},
  journal={IEEE Access}, 
  title={Applications of Machine Learning in Human Factors and Ergonomics: A Comprehensive Review of Research From the Past Decade}, 
  year={2025},
  volume={13},
  number={},
  pages={115263-115288},
  doi={10.1109/ACCESS.2025.3585773}}

@article{chen2026multimodal,
  title   = {Multimodal Physiological Assessment of Contact-Rich Physical Human--Robot Interaction Under Varying Environmental Conditions},
  author  = {Chen, Yanyi and Wang, Xi and Deng, Min},
  journal = {arXiv preprint arXiv:2606.14969},
  year    = {2026},
  doi     = {10.48550/arXiv.2606.14969}
}

@ARTICLE{asgher2020enhanced,
AUTHOR={Asgher, Umer  and Khalil, Khurram  and Khan, Muhammad Jawad  and Ahmad, Riaz  and Butt, Shahid Ikramullah  and Ayaz, Yasar  and Naseer, Noman  and Nazir, Salman },         
TITLE={Enhanced Accuracy for Multiclass Mental Workload Detection Using Long Short-Term Memory for Brain–Computer Interface},        
JOURNAL={Frontiers in Neuroscience},        
VOLUME={Volume 14 - 2020}, 
YEAR={2020},
DOI={10.3389/fnins.2020.00584},
ISSN={1662-453X}}

@article{rojas2019,
  title = {A Variational Approach to Minimum-Jerk Trajectories for Psychological Safety in Collaborative Assembly Stations},
  volume = {4},
  ISSN = {2377-3774},
  DOI = {10.1109/lra.2019.2893018},
  number = {2},
  journal = {IEEE Robotics and Automation Letters},
  publisher = {Institute of Electrical and Electronics Engineers (IEEE)},
  author = {Rojas,  Rafael A. and Garcia,  Manuel A. Ruiz and Wehrle,  Erich and Vidoni,  Renato},
  year = {2019},
  month = Apr,
  pages = {823–829}
}

@article{viviani1995minimum,
  author    = {Viviani, Paolo and Flash, Tamar},
  title     = {Minimum-jerk, two-thirds power law, and isochrony: Converging approaches to movement planning},
  journal   = {Journal of Experimental Psychology: Human Perception and Performance},
  volume    = {21},
  number    = {1},
  pages     = {32--53},
  year      = {1995},
  month     = {feb},
  publisher = {American Psychological Association ({APA})},
  doi       = {10.1037/0096-1523.21.1.32}
}

@misc{he2026constraintgroundedreinforcementlearningvariable,
      title={Constraint-Grounded Reinforcement Learning for Variable Impedance Control in Contact-Rich Robotic Insertion}, 
      author={Lin He and Min Deng},
      year={2026},
      eprint={2609.13516},
      archivePrefix={arXiv},
      primaryClass={cs.RO},
      url={https://arxiv.org/abs/2609.13516}, 
}

\end{document}